\documentclass[sigconf, nonacm]{acmart}
\usepackage{outlines}
\usepackage{hyperref}

\hypersetup{
    colorlinks=true,
    linkcolor=blue,
    filecolor=magenta,      
    urlcolor=cyan,
    pdftitle={Overleaf Example},
    pdfpagemode=FullScreen,
    }

\AtBeginDocument{%
  \providecommand\BibTeX{{%
    \normalfont B\kern-0.5em{\scshape i\kern-0.25em b}\kern-0.8em\TeX}}}

\setcopyright{acmcopyright}
\copyrightyear{2018}
\acmYear{2018}
\acmDOI{XXXXXXX.XXXXXXX}

\acmConference[Conference acronym 'XX]{Make sure to enter the correct
  conference title from your rights confirmation emai}{June 03--05,
  2018}{Woodstock, NY}
\acmPrice{15.00}
\acmISBN{978-1-4503-XXXX-X/18/06}

\begin{document}

\title{Hyperbolic Graph Representation Learning: A Tutorial}

\begin{abstract}
Graph-structured data are widespread in the real-world applications, such as social networks, recommender systems, knowledge graphs, chemical molecules etc. Despite the success of Euclidean space for graph-related learning tasks, its ability to model complex patterns is essentially constrained by its polynomially growing capacity. Recently, hyperbolic spaces have emerged as a promising alternative for processing graph data with tree-like structure or power-law distribution, owing to its exponential growth property.  Different from the Euclidean space which expands polynomially, the hyperbolic space grows exponentially which makes it gains natural advantages in abstracting tree-like or scale-free graphs with hierarchical organizations.

In this tutorial, we aim to give an introduction to this emerging field of graph representation learning, with the express purpose of being accessible to all audiences. We first give a brief introduction to graph representation learning as well as some preliminary Riemannian and hyperbolic geometry. We then comprehensively revisit the hyperbolic embedding techniques including hyperbolic shallow models and hyperbolic neural networks. In addition, we 
introduce the technical details of the current hyperbolic graph neural networks, by unifying them into a general framework and summarizing the variants of each component. Moreover, we further introduce a series of related applications in a variety of fields. In the last part, we discuss several advanced topics about hyperbolic geometry for graph representation learning, which potentially serve as guidelines for further flourishing the non-Euclidean graph learning community.


\end{abstract}

\keywords{Graph representation learning, hyperbolic space, graphs and networks}


\author{Min Zhou}

\affiliation{%
  \institution{Noah's Ark Lab, Huawei Technologies}
  \city{Shenzhen}
  \country{China}
}
\email{zhoumin27@huawei.com}

\author{Menglin Yang}

\affiliation{%
  \institution{The Chinese University of Hong Kong}
  \city{Hong Kong}
  \country{China}
}
\email{mlyang@cse.cuhk.edu.hk}

\author{Pujia Pan}
\affiliation{%
  \institution{ Noah's Ark Lab, Huawei Technologies}
  \city{Shenzhen}
  \country{China}
}
\email{panlujia@huawei.com}

\author{Irwin King}
\affiliation{%
  \institution{The Chinese University of Hong Kong}
  \city{Hong Kong}
  \country{China}
}
\email{king@cse.cuhk.edu.hk}

\maketitle

\section{Introduction}

Graph-structured data are ubiquitous in the real-world applications, ranging from social networks, recommender systems, knowledge graphs to chemical molecules. Despite the effectiveness of Euclidean space for graph-related learning tasks, its ability to encode complex patterns is intrinsically limited by its polynomially expanding capacity. Although nonlinear techniques~\cite{non_linear_embedding} assist to mitigate this issue, complex graph patterns may still need an embedding dimensionality that is computationally intractable. As revealed by recent research~\cite{bronstein2017geometric} that many complex data shows non-Euclidean underlying anatomy, for example, the datasets with tree-like structure (e.g., hierarchies, power-law distribution) extensively exists in many real-world networks, such as the hypernym structure in natural languages, the subordinate structure of entities in the knowledge graph, the organizational structure for financial fraud, and item-user interactions in recommender systems.
In these situations, Euclidean space fails to produce the most powerful or adequate geometrical representations.

Recently, hyperbolic space has gained increasing popularity in the representation learning community~\cite{hgcn2019,liu2019HGNN,peng2021hyperbolic,liu2022enhancing,yang2021hyper,chen2021modeling} and various applications~\cite{sun2021hgcf,yang2021discrete,yang2022hicf,yang2022hrcf}. 
The typical geometric property of hyperbolic space is that its volume increases exponentially in proportion to its radius, whereas the Euclidean space grows polynomially. 
Such a geometric trait brings two benefits enabling it to well deal with the complex real-world scenarios. The \textit{first} one is that hyperbolic space exhibits minimal distortion and fits the hierarchies particularly well since the space closely matches the growth rate of tree-like data while the Euclidean space cannot. The \textit{second} is that even though with a low dimensional embedding space, hyperbolic models are surprisingly able to produce high-quality representation, which makes it to be especially favorable in low-memory and low-storage scenarios.

We believe the nascent topic highly fits the interest of the machine learning and data mining community. However, as far as we know, there are few related tutorials or seminars at present.  The most related one is a workshop jointly held in Neurips 2020\footnote{https://sites.google.com/view/diffgeo4dl/home}, which gives a general overview of applications of differential geometry for deep learning without the emphasis on the graph-structured situations and hyperbolic geometry. Hence, it is necessary to have this tutorial, providing a more comprehensive discussion of the methods,applications, and challenges of this fast growing area.

\section{Target Audience and Requirements}
The tutorial targets machine learning researchers of any background. The young researchers, who are interested in the domain of graph representation learning, non-Euclidean representation learning, network embedding, and on how hyperbolic geometry can be applied in multiple scenarios, are especially welcome. No special background on data mining is required for the participants but it is helpful if the audiences already have some knowledge about concepts from mathematics, such as vector spaces and manifolds. If not, we  also provide intuitive descriptions for all of them during the tutorial. The expected audience size is 200. No special technical equipment is needed and some way of taking notes is suggested.

\section{Outline}
The tutorial is mainly organized and refers to the recently released survey papers \cite{peng2021hyperbolic} and \cite{yang2022hyperbolic} but with more emphasis on the challenges by presenting the potential or newly developed solutions to address them. The outline is sketched as following:
\begin{outline}
\1 Introduction (Min, \textbf{30 min}) \cite{gcn2017,2010hyperbolic,zhou2020graph}
    \2 An overview of graph representation learning 
    \2 Brief introduction of Riemannian geometry  
    \2 The motivation hyperbolic graph representation learning 
    
\1 Hyperbolic graph representation learning (HGRL) (Menglin, \textbf{30 min}) \cite{yang2022hyperbolic,peng2021hyperbolic}
    \2 Hyperbolic shallow models 
    \2 Hyperbolic neural networks 
    \3 Hyperbolic MLR
    \3 Hyperbolic RNN
  
    \2 Hyperbolic Graph Neural Networks 
    \3 Hyperbolic feature transformation
    \3 Hyperbolic neighborhood aggregation
    \3 Hyperbolic non-linear activation
\1 Applications (Menglin, \textbf{45 min}) 
    \2 HGRL for recommender systems \cite{HyperML2020,sun2021hgcf,yang2022hrcf,chen2021modeling,wang2021fully} 
    \2 HGRL for knowledge graph  \cite{wu2021hyperbolic,bose2020latent} 
    \2 HGRL for other applications \cite{yang2021discrete,sawhney2021exploring} 
   
\1 Advanced Topics (Min, \textbf{60 min})
    \2 Complex structures ~\cite{zhu2020graph,bachmann2020constant,wang2021mixed,fu2021ace,li2022curvature} 
    \2 Evolving interactions~\cite{sun2021hyperbolic,yang2021discrete} 
    \2 Geometry-aware Learning ~\cite{yang2022hrcf} 
    
    \2 Trustworthy and scalability  ~\cite{zhang2021we,suzuki2021GraphEmbedding,chen2021fully,sun2021hgcf}

\end{outline}
Video records and the slides are now accessed via the tutorial homepage(\url{https://hyperbolicgraphlearning.github.io/} )for readers' convenience. 

\section{Contributors}
\begin{itemize}
    \item \textbf{Min Zhou} is currently a Principal Research Engineer of Huawei Noah’s ARK LAB, Shenzhen, China. She received the B.S. degree in Automation from the University of Science and Technology of China in 2012, and the Ph.D. degree from Industrial Systems Engineering and Management Department, National University of Singapore in 2016, respectively. Her interests include pattern mining and machine learning, and their applications in sequence and graph data.  She has published several works related to graph learning and mining on top conferences and journals.
    
    \item \textbf{Menglin Yang} is currently a final-year PhD student in the Department of Computer Science and Engineering, The Chinese University of Hong Kong (CUHK). His research interests include hyperbolic graph learning and machine learning. His several works related to hyperbolic graph representation learning were accepted at recent top conferences, including KDD 2021, WSDM 2022, WWW 2022, KDD2022, SIGIR2022.
    
    \item \textbf{Lujia Pan} is the expert of Huawei Noah’s Ark Lab, Shenzhen, China. She currently heads the Intelligent operation and maintenance team and is working closely with a group of researchers and engineers on different projects such as fault diagnosis, anomaly detection, prediction in ICT(information and communications technology) network. Her research interests are on various issues related to improving the performance and reliability of intelligent operation and maintenance, including representation learning, time series analysis, label denoising, and active learning. In these research areas, she has published more than 20 technical papers in journals and conferences and is the inventor of more than 40 patents. She is currently a part-time PhD student in Department of Control Systems and Engineering, Xi’an Jiaotong University. Before that she received her B.S. and M.S. degrees in information engineering from Chongqing University of Posts and Telecommunications.
    
    \item \textbf{Irwin King} is the Chair and Professor of Computer Science and Engineering at The Chinese University of Hong Kong. His research interests include machine learning, social computing, AI, web intelligence, data mining, and multimedia information processing. In these research areas, he has over 300 technical publications in journals and conferences. He is an Associate Editor of the Journal of Neural Networks (NN). He is an IEEE Fellow, an ACM Distinguished Member, and a Fellow of Hong Kong Institute of Engineers (HKIE). He has served as the President of the International Neural Network Society (INNS), General Co-chair of The WebConf 2020, ICONIP 2020, WSDM 2011, RecSys 2013, ACML 2015, and in various capacities in a number of top conferences and societies such as WWW, NIPS, ICML, IJCAI, AAAI, APNNS, etc. He is the recipient of the ACM CIKM 2019 Test of Time Award, the ACM SIGIR 2020 Test of Time Award, and 2020 APNNS Outstanding Achievement Award for his contributions made in social computing with machine learning. In early 2010 while on leave with AT\&T Labs Research, San Francisco, he taught classes as a Visiting Professor at UC Berkeley. He received his B.Sc. degree in Engineering and Applied Science from California Institute of Technology (Caltech), Pasadena and his M.Sc. and Ph.D. degree in Computer Science from the University of Southern California (USC), Los Angeles.
    
\end{itemize}

\section{Societal Impact}

In this tutorial, we focus on the emerging field of graph representation learning.  We identify several challenges and present the potential solutions to address them, including some attempts conducted on our own. Our tutorial could share the main ideas among researchers, with the aim of pushing the boundary of graph representation research. Meanwhile, we have involved the contents of the applications in various domains. 
For instance, The hyperbolic graph representation techniques have been applied to understand cell developmental processes~\cite{klimovskaia2020poincare}, which helps to discover hierarchies from scRNAseq data. They have also been applied in disease spreading analysis~\cite{hgcn2019} and drug discovery such as molecular property prediction~\cite{yu2020semi}. These lines of research no doubt help chemists to discover the new drugs more efficiently, encouraging large positive potential social impacts.  We believe that the more we understand hyperbolic space powered graph representation techniques,  the better we can solve these challenging problems and further benefit society.

\bibliographystyle{ACM-Reference-Format}
\bibliography{reference}


\begin{thebibliography}{32}


\ifx \showCODEN    \undefined \def \showCODEN     #1{\unskip}     \fi
\ifx \showDOI      \undefined \def \showDOI       #1{#1}\fi
\ifx \showISBNx    \undefined \def \showISBNx     #1{\unskip}     \fi
\ifx \showISBNxiii \undefined \def \showISBNxiii  #1{\unskip}     \fi
\ifx \showISSN     \undefined \def \showISSN      #1{\unskip}     \fi
\ifx \showLCCN     \undefined \def \showLCCN      #1{\unskip}     \fi
\ifx \shownote     \undefined \def \shownote      #1{#1}          \fi
\ifx \showarticletitle \undefined \def \showarticletitle #1{#1}   \fi
\ifx \showURL      \undefined \def \showURL       {\relax}        \fi
\providecommand\bibfield[2]{#2}
\providecommand\bibinfo[2]{#2}
\providecommand\natexlab[1]{#1}
\providecommand\showeprint[2][]{arXiv:#2}

\bibitem[Bachmann et~al\mbox{.}(2020)]%
        {bachmann2020constant}
\bibfield{author}{\bibinfo{person}{Gregor Bachmann}, \bibinfo{person}{Gary
  B{\'e}cigneul}, {and} \bibinfo{person}{Octavian Ganea}.}
  \bibinfo{year}{2020}\natexlab{}.
\newblock \showarticletitle{Constant curvature graph convolutional networks}.
  In \bibinfo{booktitle}{\emph{ICML}}. PMLR, \bibinfo{pages}{486--496}.
\newblock


\bibitem[Bose et~al\mbox{.}(2020)]%
        {bose2020latent}
\bibfield{author}{\bibinfo{person}{Avishek~Joey Bose}, \bibinfo{person}{Ariella
  Smofsky}, \bibinfo{person}{Renjie Liao}, \bibinfo{person}{Prakash
  Panangaden}, {and} \bibinfo{person}{William~L Hamilton}.}
  \bibinfo{year}{2020}\natexlab{}.
\newblock \showarticletitle{Latent Variable Modelling with Hyperbolic
  Normalizing Flows}. In \bibinfo{booktitle}{\emph{ICML}}.
  \bibinfo{pages}{1045--1055}.
\newblock


\bibitem[Bouchard et~al\mbox{.}(2015)]%
        {non_linear_embedding}
\bibfield{author}{\bibinfo{person}{Guillaume Bouchard}, \bibinfo{person}{Sameer
  Singh}, {and} \bibinfo{person}{Theo Trouillon}.}
  \bibinfo{year}{2015}\natexlab{}.
\newblock \showarticletitle{On Approximate Reasoning Capabilities of Low-Rank
  Vector Spaces.}. In \bibinfo{booktitle}{\emph{AAAI spring symposia}}.
\newblock


\bibitem[Bronstein et~al\mbox{.}(2017)]%
        {bronstein2017geometric}
\bibfield{author}{\bibinfo{person}{Michael~M Bronstein}, \bibinfo{person}{Joan
  Bruna}, \bibinfo{person}{Yann LeCun}, \bibinfo{person}{Arthur Szlam}, {and}
  \bibinfo{person}{Pierre Vandergheynst}.} \bibinfo{year}{2017}\natexlab{}.
\newblock \showarticletitle{Geometric deep learning: going beyond euclidean
  data}.
\newblock \bibinfo{journal}{\emph{IEEE Signal Processing Magazine}}
  \bibinfo{volume}{34}, \bibinfo{number}{4} (\bibinfo{year}{2017}),
  \bibinfo{pages}{18--42}.
\newblock


\bibitem[Chami et~al\mbox{.}(2019)]%
        {hgcn2019}
\bibfield{author}{\bibinfo{person}{Ines Chami}, \bibinfo{person}{Zhitao Ying},
  \bibinfo{person}{Christopher R{\'e}}, {and} \bibinfo{person}{Jure Leskovec}.}
  \bibinfo{year}{2019}\natexlab{}.
\newblock \showarticletitle{Hyperbolic graph convolutional neural networks}. In
  \bibinfo{booktitle}{\emph{NeurIPS}}. \bibinfo{pages}{4868--4879}.
\newblock


\bibitem[Chen et~al\mbox{.}(2021)]%
        {chen2021fully}
\bibfield{author}{\bibinfo{person}{Weize Chen}, \bibinfo{person}{Xu Han},
  \bibinfo{person}{Yankai Lin}, \bibinfo{person}{Hexu Zhao},
  \bibinfo{person}{Zhiyuan Liu}, \bibinfo{person}{Peng Li},
  \bibinfo{person}{Maosong Sun}, {and} \bibinfo{person}{Jie Zhou}.}
  \bibinfo{year}{2021}\natexlab{}.
\newblock \showarticletitle{Fully Hyperbolic Neural Networks}.
\newblock \bibinfo{journal}{\emph{arXiv preprint arXiv:2105.14686}}
  (\bibinfo{year}{2021}).
\newblock


\bibitem[Chen et~al\mbox{.}(2022)]%
        {chen2021modeling}
\bibfield{author}{\bibinfo{person}{Yankai Chen}, \bibinfo{person}{Menglin
  Yang}, \bibinfo{person}{Yingxue Zhang}, \bibinfo{person}{Mengchen Zhao},
  \bibinfo{person}{Ziqiao Meng}, \bibinfo{person}{Jianye Hao}, {and}
  \bibinfo{person}{Irwin King}.} \bibinfo{year}{2022}\natexlab{}.
\newblock \showarticletitle{Modeling Scale-free Graphs for Knowledge-aware
  Recommendation}.
\newblock  (\bibinfo{year}{2022}).
\newblock


\bibitem[Fu et~al\mbox{.}(2021)]%
        {fu2021ace}
\bibfield{author}{\bibinfo{person}{Xingcheng Fu}, \bibinfo{person}{Jianxin Li},
  \bibinfo{person}{Jia Wu}, \bibinfo{person}{Qingyun Sun},
  \bibinfo{person}{Cheng Ji}, \bibinfo{person}{Senzhang Wang},
  \bibinfo{person}{Jiajun Tan}, \bibinfo{person}{Hao Peng}, {and}
  \bibinfo{person}{S~Yu Philip}.} \bibinfo{year}{2021}\natexlab{}.
\newblock \showarticletitle{{ACE-HGNN}: Adaptive Curvature Exploration
  Hyperbolic Graph Neural Network}. In \bibinfo{booktitle}{\emph{ICDM}}.
  \bibinfo{pages}{111--120}.
\newblock


\bibitem[Kipf and Welling(2017)]%
        {gcn2017}
\bibfield{author}{\bibinfo{person}{Thomas~N Kipf} {and} \bibinfo{person}{Max
  Welling}.} \bibinfo{year}{2017}\natexlab{}.
\newblock \showarticletitle{Semi-Supervised Classification with Graph
  Convolutional Networks}. In \bibinfo{booktitle}{\emph{ICLR}}.
\newblock


\bibitem[Klimovskaia et~al\mbox{.}(2020)]%
        {klimovskaia2020poincare}
\bibfield{author}{\bibinfo{person}{Anna Klimovskaia}, \bibinfo{person}{David
  Lopez-Paz}, \bibinfo{person}{L{\'e}on Bottou}, {and}
  \bibinfo{person}{Maximilian Nickel}.} \bibinfo{year}{2020}\natexlab{}.
\newblock \showarticletitle{Poincar{\'e} maps for analyzing complex hierarchies
  in single-cell data}.
\newblock \bibinfo{journal}{\emph{Nature Communications}} \bibinfo{volume}{11},
  \bibinfo{number}{1} (\bibinfo{year}{2020}), \bibinfo{pages}{1--9}.
\newblock


\bibitem[Krioukov et~al\mbox{.}(2010)]%
        {2010hyperbolic}
\bibfield{author}{\bibinfo{person}{Dmitri Krioukov},
  \bibinfo{person}{Fragkiskos Papadopoulos}, \bibinfo{person}{Maksim Kitsak},
  \bibinfo{person}{Amin Vahdat}, {and} \bibinfo{person}{Mari{\'a}n
  Bogun{\'a}}.} \bibinfo{year}{2010}\natexlab{}.
\newblock \showarticletitle{Hyperbolic geometry of complex networks}.
\newblock \bibinfo{journal}{\emph{Physical Review E}} \bibinfo{volume}{82},
  \bibinfo{number}{3} (\bibinfo{year}{2010}), \bibinfo{pages}{036106}.
\newblock


\bibitem[Li et~al\mbox{.}(2022)]%
        {li2022curvature}
\bibfield{author}{\bibinfo{person}{Jianxin Li}, \bibinfo{person}{Xingcheng Fu},
  \bibinfo{person}{Qingyun Sun}, \bibinfo{person}{Cheng Ji},
  \bibinfo{person}{Jiajun Tan}, \bibinfo{person}{Jia Wu}, {and}
  \bibinfo{person}{Hao Peng}.} \bibinfo{year}{2022}\natexlab{}.
\newblock \showarticletitle{Curvature Graph Generative Adversarial Networks}.
\newblock \bibinfo{journal}{\emph{arXiv preprint arXiv:2203.01604}}
  (\bibinfo{year}{2022}).
\newblock


\bibitem[Liu et~al\mbox{.}(2022)]%
        {liu2022enhancing}
\bibfield{author}{\bibinfo{person}{Jiahong Liu}, \bibinfo{person}{Menglin
  Yang}, \bibinfo{person}{Min Zhou}, \bibinfo{person}{Shanshan Feng}, {and}
  \bibinfo{person}{Philippe Fournier-Viger}.} \bibinfo{year}{2022}\natexlab{}.
\newblock \showarticletitle{Enhancing Hyperbolic Graph Embeddings via
  Contrastive Learning}. In \bibinfo{booktitle}{\emph{NeurIPS 2nd SSL
  workshop}}.
\newblock


\bibitem[Liu et~al\mbox{.}(2019)]%
        {liu2019HGNN}
\bibfield{author}{\bibinfo{person}{Qi Liu}, \bibinfo{person}{Maximilian
  Nickel}, {and} \bibinfo{person}{Douwe Kiela}.}
  \bibinfo{year}{2019}\natexlab{}.
\newblock \showarticletitle{Hyperbolic graph neural networks}. In
  \bibinfo{booktitle}{\emph{NeurIPS}}. \bibinfo{pages}{8230--8241}.
\newblock


\bibitem[Peng et~al\mbox{.}(2021)]%
        {peng2021hyperbolic}
\bibfield{author}{\bibinfo{person}{Wei Peng}, \bibinfo{person}{Tuomas Varanka},
  \bibinfo{person}{Abdelrahman Mostafa}, \bibinfo{person}{Henglin Shi}, {and}
  \bibinfo{person}{Guoying Zhao}.} \bibinfo{year}{2021}\natexlab{}.
\newblock \showarticletitle{Hyperbolic deep neural networks: A survey}.
\newblock \bibinfo{journal}{\emph{TPAMI}} (\bibinfo{year}{2021}).
\newblock


\bibitem[Sawhney et~al\mbox{.}(2021)]%
        {sawhney2021exploring}
\bibfield{author}{\bibinfo{person}{Ramit Sawhney}, \bibinfo{person}{Shivam
  Agarwal}, \bibinfo{person}{Arnav Wadhwa}, {and} \bibinfo{person}{Rajiv
  Shah}.} \bibinfo{year}{2021}\natexlab{}.
\newblock \showarticletitle{Exploring the scale-free nature of stock markets:
  Hyperbolic graph learning for algorithmic trading}. In
  \bibinfo{booktitle}{\emph{WWW}}. \bibinfo{pages}{11--22}.
\newblock


\bibitem[Sun et~al\mbox{.}(2021a)]%
        {sun2021hgcf}
\bibfield{author}{\bibinfo{person}{Jianing Sun}, \bibinfo{person}{Zhaoyue
  Cheng}, \bibinfo{person}{Saba Zuberi}, \bibinfo{person}{Felipe P{\'e}rez},
  {and} \bibinfo{person}{Maksims Volkovs}.} \bibinfo{year}{2021}\natexlab{a}.
\newblock \showarticletitle{{HGCF}: Hyperbolic Graph Convolution Networks for
  Collaborative Filtering}. In \bibinfo{booktitle}{\emph{WWW}}.
  \bibinfo{pages}{593--601}.
\newblock


\bibitem[Sun et~al\mbox{.}(2021b)]%
        {sun2021hyperbolic}
\bibfield{author}{\bibinfo{person}{Li Sun}, \bibinfo{person}{Zhongbao Zhang},
  \bibinfo{person}{Jiawei Zhang}, \bibinfo{person}{Feiyang Wang},
  \bibinfo{person}{Hao Peng}, \bibinfo{person}{Sen Su}, {and}
  \bibinfo{person}{Philip~S Yu}.} \bibinfo{year}{2021}\natexlab{b}.
\newblock \showarticletitle{Hyperbolic variational graph neural network for
  modeling dynamic graphs}. In \bibinfo{booktitle}{\emph{AAAI}},
  Vol.~\bibinfo{volume}{35}. \bibinfo{pages}{4375--4383}.
\newblock


\bibitem[Suzuki et~al\mbox{.}(2021)]%
        {suzuki2021GraphEmbedding}
\bibfield{author}{\bibinfo{person}{Atsushi Suzuki}, \bibinfo{person}{Atsushi
  Nitanda}, \bibinfo{person}{Linchuan Xu}, \bibinfo{person}{Kenji Yamanishi},
  \bibinfo{person}{Marc Cavazza}, {et~al\mbox{.}}}
  \bibinfo{year}{2021}\natexlab{}.
\newblock \showarticletitle{Generalization Bounds for Graph Embedding Using
  Negative Sampling: Linear vs Hyperbolic}. In
  \bibinfo{booktitle}{\emph{NeurIPS}}.
\newblock


\bibitem[Vinh~Tran et~al\mbox{.}(2020)]%
        {HyperML2020}
\bibfield{author}{\bibinfo{person}{Lucas Vinh~Tran}, \bibinfo{person}{Yi Tay},
  \bibinfo{person}{Shuai Zhang}, \bibinfo{person}{Gao Cong}, {and}
  \bibinfo{person}{Xiaoli Li}.} \bibinfo{year}{2020}\natexlab{}.
\newblock \showarticletitle{Hyper{ML}: A Boosting Metric Learning Approach in
  Hyperbolic Space for Recommender Systems}. In
  \bibinfo{booktitle}{\emph{WSDM}}. \bibinfo{address}{New York, NY, USA},
  \bibinfo{pages}{609–617}.
\newblock


\bibitem[Wang et~al\mbox{.}(2021a)]%
        {wang2021fully}
\bibfield{author}{\bibinfo{person}{Liping Wang}, \bibinfo{person}{Fenyu Hu},
  \bibinfo{person}{Shu Wu}, {and} \bibinfo{person}{Liang Wang}.}
  \bibinfo{year}{2021}\natexlab{a}.
\newblock \showarticletitle{Fully Hyperbolic Graph Convolution Network for
  Recommendation}.
\newblock \bibinfo{journal}{\emph{arXiv preprint arXiv:2108.04607}}
  (\bibinfo{year}{2021}).
\newblock


\bibitem[Wang et~al\mbox{.}(2021b)]%
        {wang2021mixed}
\bibfield{author}{\bibinfo{person}{Shen Wang}, \bibinfo{person}{Xiaokai Wei},
  \bibinfo{person}{Cicero~Nogueira Nogueira~dos Santos},
  \bibinfo{person}{Zhiguo Wang}, \bibinfo{person}{Ramesh Nallapati},
  \bibinfo{person}{Andrew Arnold}, \bibinfo{person}{Bing Xiang},
  \bibinfo{person}{Philip~S Yu}, {and} \bibinfo{person}{Isabel~F Cruz}.}
  \bibinfo{year}{2021}\natexlab{b}.
\newblock \showarticletitle{Mixed-curvature multi-relational graph neural
  network for knowledge graph completion}. In \bibinfo{booktitle}{\emph{WWW}}.
  \bibinfo{pages}{1761--1771}.
\newblock


\bibitem[Wu et~al\mbox{.}(2021)]%
        {wu2021hyperbolic}
\bibfield{author}{\bibinfo{person}{Zhenxing Wu}, \bibinfo{person}{Dejun Jiang},
  \bibinfo{person}{Chang-Yu Hsieh}, \bibinfo{person}{Guangyong Chen},
  \bibinfo{person}{Ben Liao}, \bibinfo{person}{Dongsheng Cao}, {and}
  \bibinfo{person}{Tingjun Hou}.} \bibinfo{year}{2021}\natexlab{}.
\newblock \showarticletitle{Hyperbolic relational graph convolution networks
  plus: a simple but highly efficient QSAR-modeling method}.
\newblock \bibinfo{journal}{\emph{Briefings in Bioinformatics}}
  \bibinfo{volume}{22}, \bibinfo{number}{5} (\bibinfo{year}{2021}),
  \bibinfo{pages}{bbab112}.
\newblock


\bibitem[Yang et~al\mbox{.}(2021a)]%
        {yang2021hyper}
\bibfield{author}{\bibinfo{person}{Haoran Yang}, \bibinfo{person}{Hongxu Chen},
  \bibinfo{person}{Lin Li}, \bibinfo{person}{Philip~S Yu}, {and}
  \bibinfo{person}{Guandong Xu}.} \bibinfo{year}{2021}\natexlab{a}.
\newblock \showarticletitle{Hyper Meta-Path Contrastive Learning for
  Multi-Behavior Recommendation}.
\newblock \bibinfo{journal}{\emph{arXiv preprint arXiv:2109.02859}}
  (\bibinfo{year}{2021}).
\newblock


\bibitem[Yang et~al\mbox{.}(2022a)]%
        {yang2022hicf}
\bibfield{author}{\bibinfo{person}{Menglin Yang}, \bibinfo{person}{Zhihao Li},
  \bibinfo{person}{Min Zhou}, \bibinfo{person}{Jiahong Liu}, {and}
  \bibinfo{person}{Irwin King}.} \bibinfo{year}{2022}\natexlab{a}.
\newblock \showarticletitle{HICF: Hyperbolic Informative Collaborative
  Filtering}. In \bibinfo{booktitle}{\emph{Proceedings of the 28th ACM SIGKDD
  Conference on Knowledge Discovery and Data Mining}}.
  \bibinfo{pages}{2212--2221}.
\newblock


\bibitem[Yang et~al\mbox{.}(2021b)]%
        {yang2021discrete}
\bibfield{author}{\bibinfo{person}{Menglin Yang}, \bibinfo{person}{Min Zhou},
  \bibinfo{person}{Marcus Kalander}, \bibinfo{person}{Zengfeng Huang}, {and}
  \bibinfo{person}{Irwin King}.} \bibinfo{year}{2021}\natexlab{b}.
\newblock \showarticletitle{Discrete-time Temporal Network Embedding via
  Implicit Hierarchical Learning in Hyperbolic Space}. In
  \bibinfo{booktitle}{\emph{KDD}}. \bibinfo{pages}{1975--1985}.
\newblock


\bibitem[Yang et~al\mbox{.}(2022b)]%
        {yang2022hyperbolic}
\bibfield{author}{\bibinfo{person}{Menglin Yang}, \bibinfo{person}{Min Zhou},
  \bibinfo{person}{Zhihao Li}, \bibinfo{person}{Jiahong Liu},
  \bibinfo{person}{Lujia Pan}, \bibinfo{person}{Hui Xiong}, {and}
  \bibinfo{person}{Irwin King}.} \bibinfo{year}{2022}\natexlab{b}.
\newblock \showarticletitle{Hyperbolic Graph Neural Networks: A Review of
  Methods and Applications}.
\newblock \bibinfo{journal}{\emph{arXiv e-prints}} (\bibinfo{year}{2022}),
  \bibinfo{pages}{arXiv 2202.13852}.
\newblock


\bibitem[Yang et~al\mbox{.}(2022c)]%
        {yang2022hrcf}
\bibfield{author}{\bibinfo{person}{Menglin Yang}, \bibinfo{person}{Min Zhou},
  \bibinfo{person}{Jiahong Liu}, \bibinfo{person}{Defu Lian}, {and}
  \bibinfo{person}{Irwin King}.} \bibinfo{year}{2022}\natexlab{c}.
\newblock \showarticletitle{{HRCF}: Enhancing Collaborative Filtering via
  Hyperbolic Geometric Regularization}. In \bibinfo{booktitle}{\emph{WWW}}.
\newblock


\bibitem[Yu et~al\mbox{.}(2020)]%
        {yu2020semi}
\bibfield{author}{\bibinfo{person}{Ke Yu}, \bibinfo{person}{Shyam Visweswaran},
  {and} \bibinfo{person}{Kayhan Batmanghelich}.}
  \bibinfo{year}{2020}\natexlab{}.
\newblock \showarticletitle{Semi-supervised hierarchical drug embedding in
  hyperbolic space}.
\newblock \bibinfo{journal}{\emph{Journal of chemical information and
  modeling}} \bibinfo{volume}{60}, \bibinfo{number}{12} (\bibinfo{year}{2020}),
  \bibinfo{pages}{5647--5657}.
\newblock


\bibitem[Zhang et~al\mbox{.}(2021)]%
        {zhang2021we}
\bibfield{author}{\bibinfo{person}{Sixiao Zhang}, \bibinfo{person}{Hongxu
  Chen}, \bibinfo{person}{Xiao Ming}, \bibinfo{person}{Lizhen Cui},
  \bibinfo{person}{Hongzhi Yin}, {and} \bibinfo{person}{Guandong Xu}.}
  \bibinfo{year}{2021}\natexlab{}.
\newblock \showarticletitle{Where are we in embedding spaces? A Comprehensive
  Analysis on Network Embedding Approaches for Recommender Systems}. In
  \bibinfo{booktitle}{\emph{KDD}}.
\newblock


\bibitem[Zhou et~al\mbox{.}(2020)]%
        {zhou2020graph}
\bibfield{author}{\bibinfo{person}{Jie Zhou}, \bibinfo{person}{Ganqu Cui},
  \bibinfo{person}{Shengding Hu}, \bibinfo{person}{Zhengyan Zhang},
  \bibinfo{person}{Cheng Yang}, \bibinfo{person}{Zhiyuan Liu},
  \bibinfo{person}{Lifeng Wang}, \bibinfo{person}{Changcheng Li}, {and}
  \bibinfo{person}{Maosong Sun}.} \bibinfo{year}{2020}\natexlab{}.
\newblock \showarticletitle{Graph neural networks: A review of methods and
  applications}.
\newblock \bibinfo{journal}{\emph{AI Open}}  \bibinfo{volume}{1}
  (\bibinfo{year}{2020}), \bibinfo{pages}{57--81}.
\newblock


\bibitem[Zhu et~al\mbox{.}(2020)]%
        {zhu2020graph}
\bibfield{author}{\bibinfo{person}{Shichao Zhu}, \bibinfo{person}{Shirui Pan},
  \bibinfo{person}{Chuan Zhou}, \bibinfo{person}{Jia Wu},
  \bibinfo{person}{Yanan Cao}, {and} \bibinfo{person}{Bin Wang}.}
  \bibinfo{year}{2020}\natexlab{}.
\newblock \showarticletitle{Graph geometry interaction learning}.
\newblock \bibinfo{journal}{\emph{NeurIPS}}  \bibinfo{volume}{33}
  (\bibinfo{year}{2020}), \bibinfo{pages}{7548--7558}.
\newblock


\end{thebibliography}

\end{document}